\documentclass{article}

\usepackage{arxiv}

\usepackage[utf8]{inputenc} 
\usepackage[T1]{fontenc}    
\usepackage{hyperref}       
\usepackage{url}            
\usepackage{booktabs}       
\usepackage{amsfonts}       
\usepackage{nicefrac}       
\usepackage{microtype}      
\usepackage{lipsum}
\usepackage{graphicx}
\usepackage{subcaption}
\usepackage{amsmath}
\usepackage{float}
\usepackage{tikz}
\usetikzlibrary{shapes.geometric, arrows.meta, positioning, fit, backgrounds, calc}
\graphicspath{ {./images/} }

\title{The High Cost of Incivility: Quantifying Interaction Inefficiency via Multi-Agent Monte Carlo Simulations}

\author{
 Benedikt Mangold \\
  Technische Hochschule Nürnberg Georg Simon Ohm\\
  90489 Nuremberg, Germany\\
  \texttt{benedikt.mangold@th-nuernberg.de} \\
}

\date{December 10, 2025}

\begin{document}
\renewcommand{\shorttitle}{Quantifying Interaction Inefficiency via Multi-Agent Monte Carlo Simulations}
\maketitle
\begin{abstract}
Workplace toxicity is widely recognized as detrimental to organizational culture, yet quantifying its direct impact on operational efficiency remains methodologically challenging due to the ethical and practical difficulties of reproducing conflict in human subjects. This study leverages Large Language Model (LLM) based Multi-Agent Systems to simulate 1-on-1 adversarial debates, creating a controlled ``sociological sandbox''. We employ a Monte Carlo method to simulate hundrets of discussions, measuring the convergence time (defined as the number of arguments required to reach a conclusion) between a baseline control group and treatment groups involving agents with ``toxic'' system prompts. Our results demonstrate a statistically significant increase of approximately 25\% in the duration of conversations involving toxic participants. We propose that this ``latency of toxicity'' serves as a proxy for financial damage in corporate and academic settings. Furthermore, we demonstrate that agent-based modeling provides a reproducible, ethical alternative to human-subject research for measuring the mechanics of social friction.
\end{abstract}


\section{Introduction}
The impact of toxic behavior in professional settings is often discussed in terms of morale, psychological safety, and turnover \cite{porath2009incivility}. However, the direct \textit{inefficiency} caused by such behavior (specifically, the time lost in prolonged, circular, or unproductive communication) is difficult to isolate in the real world. Human interactions are non-reproducible; emotions cannot be ``replayed,'' and intentionally subjecting human participants to toxic behavior for the sake of measurement raises significant ethical concerns.

The advent of Large Language Models (LLMs) offers a novel solution: the use of ``Generative Agents'' \cite{park2023generative} to simulate human social dynamics. By utilizing agents with distinct personae, researchers can create a ``Multi-Agent Discussion'' (MAD) environment where variables such as behavioral traits can be strictly controlled.

In this paper, we investigate the hypothesis that toxic behavior introduces a measurable ``friction'' into communication protocols, resulting in longer convergence times. We model this as an efficiency problem: if a conversation requires more turns to resolve, it incurs a higher cost; whether in terms of token usage (for AI) or billable hours (for humans).

Crucially, the ability to measure these efficiency losses implies a underlying deterministic structure in group dynamics. If we can reliably simulate how specific behavioral traits alter the trajectory of a debate, we can leverage this same mechanism to forecast the outcome of complex social interactions before they occur. Thus, the framework transitions from a tool for cost analysis to an engine for \textit{predictive social modeling}.

Building on this premise, we lay the groundwork for advanced applications such as ``Strategic Litigation Planning'', where defense attorneys could test narrative strategies against simulated juries to predict verdict probabilities. Crucially, the underlying framework is designed for extensibility, allowing future research to scale from dyadic interactions to larger groups, incorporate diverse behavioral archetypes beyond toxicity (e.g., leadership or sycophancy), and simulate complex multi-step collaborative tasks.

Our contributions are as follows:
\begin{itemize}
    \item We introduce a Monte Carlo simulation framework for measuring debate length between LLM agents.
    \item We quantify the impact of a toxic participant, finding a $\approx$ 25\% increase in argument count before resolution.
    \item We demonstrate that agent-based modeling provides a reproducible, ethical alternative to human-subject research for measuring social friction, serving as a baseline for future high-stakes simulations like jury modeling.
\end{itemize}

\section{Related Work}

\subsection{Generative Agents and Social Simulation}
The capability of LLMs to simulate believable human behavior has been established by \cite{park2023generative}, who demonstrated that agents could form memories, relationships, and coordinate complex activities. Building on this, \cite{aher2023using} validated the use of LLMs to replicate classic social science experiments (e.g., the Ultimatum Game), arguing that these ``silicon subjects'' provide a robust proxy for human behavioral patterns. \cite{li2023camel} further introduced ``CAMEL,'' a role-playing framework where agents interact to solve tasks, highlighting the potential for autonomous cooperation. Our work extends this by focusing not on task completion success, but on the \textit{temporal efficiency} of the interaction under adversarial conditions. Our methodology builds upon the Debate-to-Write framework proposed by \cite{hu2025debatetowritepersonadrivenmultiagentframework}, specifically their approach of assigning distinct personas to agents to drive diverse argumentative behaviors.

\subsection{Consensus and Debate in Multi-Agent Systems}
Recent work has explored how agents converge on truth or consensus. \cite{du2023improving} demonstrated that multi-agent debate improves factuality and reasoning capabilities, as agents essentially error-check one another. However, these studies typically assume cooperative intent. Our research investigates the inverse scenario: the degradation of convergence speed when one agent explicitly violates cooperative norms. This parallels findings in game theory simulations with LLMs, where \cite{akata2023playing} observed that agents can exhibit varying degrees of cooperation and defection in repeated games, influencing the collective payoff.

\subsection{Measuring Toxicity and Bias}
While substantial research focuses on benchmarks for detecting toxicity within LLM outputs, such as RealToxicityPrompts \cite{gehman2020realtoxicityprompts}, or categorizing the taxonomy of harms \cite{weidinger2021ethical}, fewer studies utilize LLMs to simulate the \textit{operational effect} of toxicity on a system's efficiency. Measuring the downstream impact of malicious behavior (rather than just its presence) is crucial for designing robust multi-agent systems and understanding human organizational dynamics.

\section{Methodology}

To isolate the effect of toxic behavior on conversation efficiency, we designed a controlled experiment using the Multi-Agent Discussion (MAD) framework.

\subsection{Experimental Setup}
\begin{figure}[H]
  \centering
  \includegraphics[width=.7\textwidth]{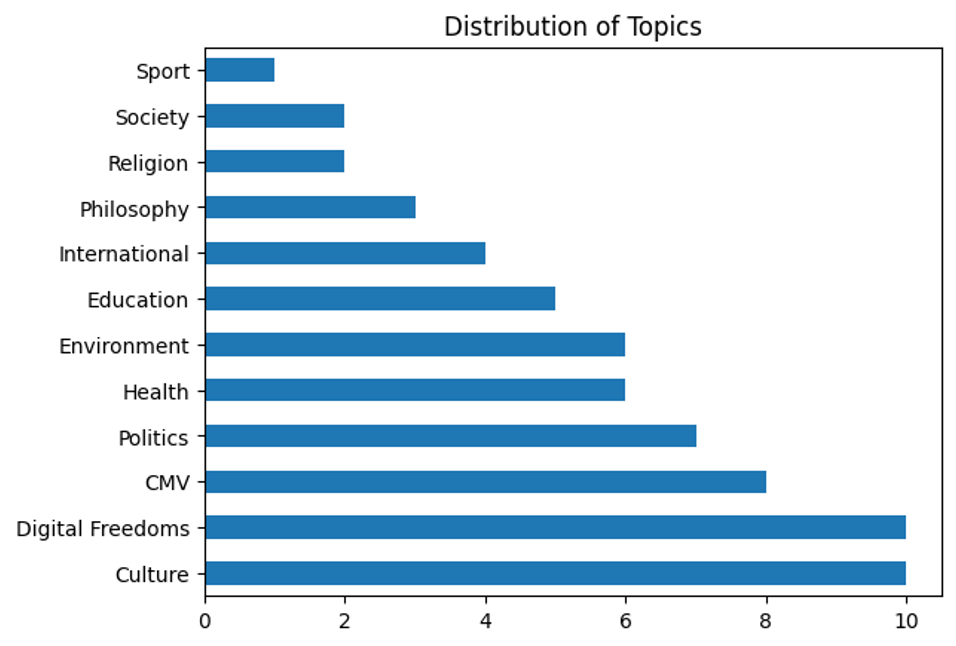}
  \caption{Amount of topics per domain (\url{https://idebate.net}), from which the debates are randomly chosen. A list of detailed topics can be found in table \ref{tab:dist_topics}}
  \label{fig:dist_topics}
\end{figure}

The core unit of our experiment is a randomized 1-on-1 debate, as proposed by . For each simulation iteration, the setup proceeds as follows :

\begin{enumerate}
    \item \textbf{Randomized Topic Selection:} A debate topic is randomly selected from a diverse pool of controversial subjects (see figure \ref{fig:dist_topics} to ensure the generalizability of result across domains.
    \item \textbf{Stance Assignment:} Two agents are instantiated (see figure \ref{fig:prompt_persona}). They are randomly assigned opposing stances: one agent acts as the \textit{Proponent} (Pro), and the other as the \textit{Opponent} (Con).
    \item \textbf{Goal Definition:} Both agents are instructed to convince their counterpart of their assigned standpoint through argumentation (see figure \ref{fig:prompt_agent}).
\end{enumerate}

\subsection{Behavioral Variable: The Toxicity Injection}
To measure the impact of behavioral traits, we differentiate between two experimental conditions:

\begin{itemize}
    \item \textbf{Control Group (Baseline):} Both agents (Pro and Con) are assigned a standard, ``Neutral/Constructive'' system prompt. They argue firmly but adhere to standard cooperative conversational norms.
    \item \textbf{Treatment Group (Toxic):} One of the two agents is randomly selected to receive the ``Toxic'' system prompt modification. This selection is independent of their stance (Pro/Con). The toxic agent is instructed to exhibit toxic behavior (as described in table \ref{tab:toxic}), while the other agent remains ``Neutral.'' (see figure \ref{fig:prompt_toxic}).
\end{itemize}

\begin{table}[H]
 \caption{Levels of toxicity and description of the behavior.}
  \centering
  \begin{tabular}{p{2cm}|p{5cm}|p{7cm}}
    \toprule
    Name     & Description    & Behaviour \\
    \midrule
    \textsf{mild} & Passive-aggressive, sarcastic, smug
& Belittles others indirectly,  implies superiority  \\
    \textsf{moderate}      & Condescending, belittling, rude
 & Dismisses others' arguments as idiotic or irrelevant,  questions their intelligence or logic
      \\
    \textsf{heavy}      & Aggressive, hostile, cruel & Insults others, uses inflammatory language, shows contempt for opposing agents\\
    \bottomrule
  \end{tabular}
  \label{tab:toxic}
\end{table}

The simulation environment was constructed to allow for autonomous interaction. After an opening statement of each agent, each sequence of arguments is executed:
\begin{itemize}
    \item An Agent is randomly chosen, and given the debates history
    \begin{itemize}
        \item tries to find the best next argument to convince the other agent
        \item acknowledge the other agent's argument and reply with the word ``convinced"
    \end{itemize} 
    \item The other Agent can react to the new argument (given the debates history) or reply with the word ``convinced".
    \item After each sequence, an external ``Moderator'' Agent is being consulted (see figure \ref{fig:prompt_moderator}) to to evaluate if the discussion is in alignment or if a conclusion has been reached.
\end{itemize}

\begin{figure}[H]
    \centering
    \includegraphics[width=0.6\linewidth]{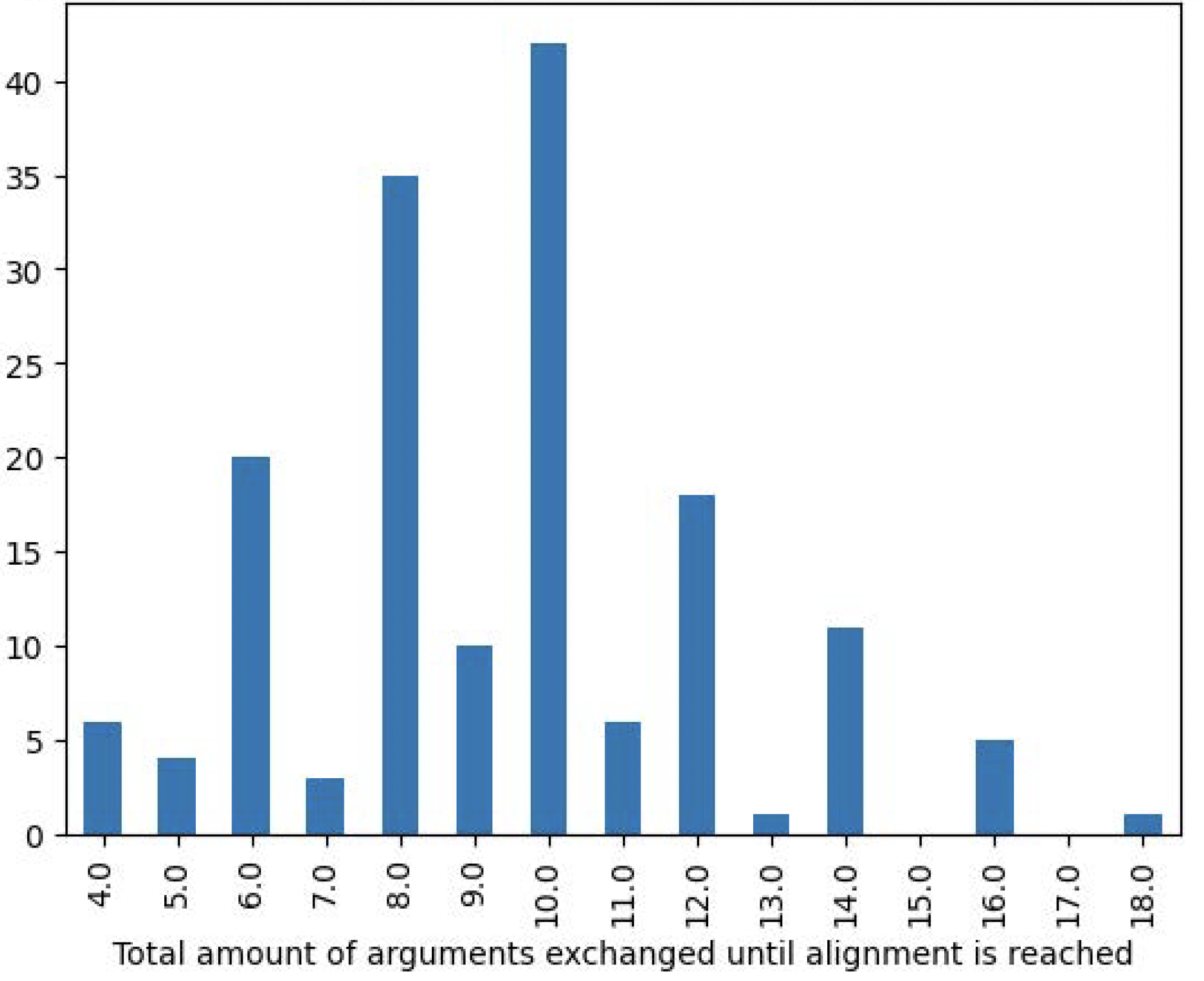}
    \caption{Arguments required until alignment without toxic behaviour (toxicity level \textsf{no}). $N=162$ debates out of a pool of 64 debates from figure \ref{fig:dist_topics}}
    \label{fig:result_no}
\end{figure}

Note that consulting the ``Moderator'' Agent is necessary as the two debating Agents may ignore the instructions to respond with ``convinced'' even though they are. The debate is considered to be \textit{ended} if either one agent replies with the word ``convinced'' or the ``Moderator'' Agent detects the discussion to be in alignment.

\subsection{Monte Carlo Simulation}
Single LLM interactions can be stochastic due to temperature settings and inherent probabilistic generation. To achieve statistical significance, we employed a Monte Carlo approach. 
\begin{itemize}
    \item \textbf{Runtime:} The simulations were conducted over a period of 3 weeks.
    \item \textbf{Iterations:} We ran up to\footnote{The higher the level of toxicity, the more likely for the Agent to refuse to create a follow-up argument which led to some failed runs} $N=162$ independent debate simulations for both control and treatment groups.
    \item \textbf{Metric:} The primary metric is $T_{\text{conv}}$, defined as the number of arguments (turns) exchanged until the conversation ends.
\end{itemize}

The flowchart in figure \ref{fig:pipeline} visualizes the complete execution pipeline, ensuring that each simulation run remains an independent, reproducible event within the Monte Carlo framework. By rigorously repeating this cycle across diverse topics and personas, we transform individual, stochastic conversation paths into robust statistical distributions. Having established this experimental apparatus, we now turn to the empirical evidence generated by these synthetic interactions. The following section analyzes these distributions to quantify the precise ``time tax'' imposed by toxic behavior on the consensus-finding process.

\begin{figure}

\resizebox{\textwidth}{\textheight}{
\begin{tikzpicture}[
    node distance=1cm and 1cm,
    font=\sffamily\small,
    >=Latex,
    startstop/.style={rectangle, rounded corners, minimum width=3cm, minimum height=1cm, text centered, draw=black!80, thick, fill=green!10},
    process/.style={rectangle, minimum width=3.5cm, minimum height=1cm, text centered, text width=3cm, draw=black!80, thick, fill=gray!5},
    decision/.style={diamond, minimum width=3cm, minimum height=1cm, text centered, text width=2cm, draw=black!80, thick, fill=yellow!10, aspect=2},
    control/.style={process, fill=cyan!10, draw=cyan!80},
    toxic/.style={process, fill=red!10, draw=red!80},
    arrow/.style={thick,->, >=Stealth},
    groupbox/.style={draw=black!40, dashed, inner sep=12pt, rounded corners, fill=white, fill opacity=0.5}
]

    \node (start) [startstop] {Start Simulation Batch};
    \node (init) [process, below=0.6cm of start] {Initialization};
    \node (topic) [process, below=0.6cm of init] {Select random topic};
    \node (stance) [process, below=0.6cm of topic] {Assign Stances:\\A (Pro) vs B (Con)};
    \node (split) [decision, below=0.6cm of stance] {Experimental Group?};

    \node (control) [control, below left=1.0cm and 0.2cm of split] {Control Group:\\Both Neutral};
    \node (toxic) [toxic, below right=1.0cm and 0.2cm of split] {Treatment Group:\\Toxic Injection};

    \node (debate) [process, below=2cm of control, xshift=-1cm] {Start Debate};
    \node (round) [process, below=0.6cm of debate] {Start Round};
    \node (pick1) [process, below=0.6cm of round] {Pick First Agent};
    \node (arg1) [process, below=0.6cm of pick1] {Convinced? If not, generate Argument};
    \node (pick2) [process, below=0.6cm of arg1] {Pick Other Agent};
    \node (arg2) [process, below=0.6cm of pick2] {Convinced? If not, generate Argument};
    \node (modcheck) [decision, below=0.8cm of arg2] {Moderator Check:\\Consensus?};

    \node (measure) [process, right=3.5cm of debate] {Measure $T_{conv}$};
    \node (store) [process, below=0.6cm of measure] {Store Argument Count};
    \node (mccheck) [decision, below=0.8cm of store] {N Iterations reached?};
    \node (result) [process, below=1.0cm of mccheck, fill=green!20, draw=green!80] {Result:\\Efficiency Gap};

    \begin{scope}[on background layer]
        \node (box1) [groupbox, fit=(init) (topic) (stance) (split) (control) (toxic), label={[anchor=north west, inner sep=5pt]north west:\textbf{1. Setup \& Randomization}}] {};
        
        \node (box2) [groupbox, fit=(debate) (round) (pick1) (arg1) (pick2) (arg2) (modcheck), label={[anchor=north west, inner sep=5pt]north west:\textbf{2. Interaction Loop}}] {};
        
        \node (box3) [groupbox, fit=(measure) (store) (mccheck) (result), label={[anchor=north west, inner sep=5pt]north west:\textbf{3. Data Collection}}] {};
    \end{scope}

    
    \draw [arrow] (start) -- (init);
    \draw [arrow] (init) -- (topic);
    \draw [arrow] (topic) -- (stance);
    \draw [arrow] (stance) -- (split);
    \draw [arrow] (split) -| node[pos=0.6, above] {Control} (control);
    \draw [arrow] (split) -| node[pos=0.6, above] {Treatment} (toxic);

    \draw [arrow] (control) -- ++(0,-1) -| (debate);
    \draw [arrow] (toxic) -- ++(0,-1) -| (debate);

    \draw [arrow] (debate) -- (round);
    \draw [arrow] (round) -- (pick1);
    \draw [arrow] (pick1) -- (arg1);
    \draw [arrow] (arg1) -- (pick2);
    \draw [arrow] (pick2) -- (arg2);
    \draw [arrow] (arg2) -- (modcheck);

    \draw [arrow] (modcheck.west) -- ++(-0.8,0) |- node[pos=0.25, anchor=east] {No (Continue)} (round.west);

    \draw [arrow] (modcheck.east) -- ++(1.5,0) |- node[pos=0.7, anchor=south] {Yes (End)} (measure.west);

    \draw [arrow] (measure) -- (store);
    \draw [arrow] (store) -- (mccheck);
    \draw [arrow] (mccheck) -- node[anchor=west] {Yes (Finish)} (result);

    \draw [arrow] (mccheck.east) -- ++(1.5,0) |- node[pos=0.25, anchor=west] {No (Repeat Batch)} (init.east);

\end{tikzpicture}
}
    \caption{Execution pipeline of the simulation study of our work.}
    \label{fig:pipeline}
\end{figure}

\section{Results}

\subsection{Convergence Latency}
Figure \ref{fig:result_no} shows the amount of arguments being exchanged across 64 different topics, $T_{\text{conv}, \text{no}}$, out of $N=162$ debates. On average, $\bar{T}_{\text{conv}, \text{no}}=9.40$ steps are required to reach an alignment between two arguing agents.

\begin{figure}[H]
\centering
\begin{subfigure}{.5\textwidth}
  \centering
  \includegraphics[width=.9\linewidth]{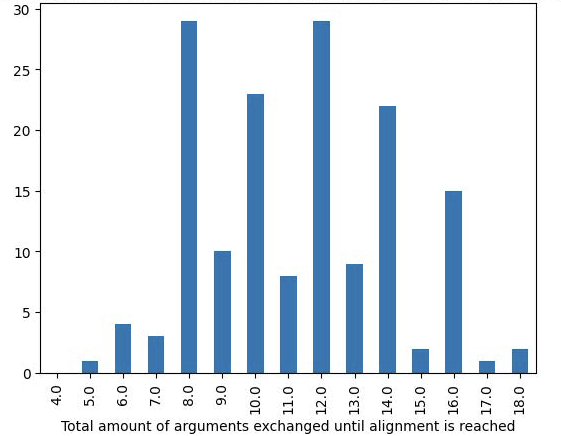}
  \caption{Toxicity level \textsf{mild} ($N=158$ debates)}
  \label{fig:result_mild}
\end{subfigure}%
\begin{subfigure}{.5\textwidth}
  \centering
  \includegraphics[width=.9\linewidth]{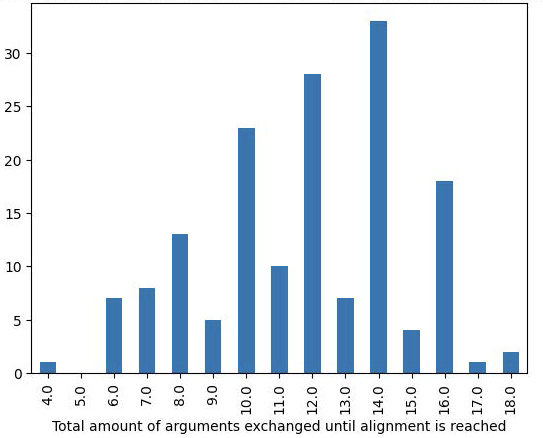}
  \caption{Toxicity level \textsf{moderate} ($N=160$ debates)}
  \label{fig:result_moderate}
\end{subfigure}
\caption{Arguments required until alignment with different levels of toxic behaviour}
\label{fig:results}
\end{figure}

The absolute frequency distribution of of $T_{\text{conv}, \text{mild}}$ and $T_{\text{conv}, \text{moderate}}$ are reported in figures \ref{fig:result_mild} and \ref{fig:result_moderate}, respectively. Simulations result in averages of $\bar{T}_{\text{conv}, \text{mild}}=11.30$ and $\bar{T}_{\text{conv}, \text{moderate}}=11.76$. Due to high refusal rates triggered by safety filters, \textsf{heavy} toxicity runs did not yield statistically sufficient valid conversations and were excluded from the efficiency analysis.

Table \ref{tab:results} compares the values of $T_{\text{conv}}$ per scenario. The results show a clear divergence between the control (\textsf{no}) and treatment (\textsf{mild} / \textsf{moderate}) groups. Discussions involving a toxic agent required significantly more steps to converge (with a $p$-value less than $1\%$).

\begin{table}[H]
 \caption{Statistics on $T_{\text{conv}}$ across control (\textsf{no}) and treatment (\textsf{mild} / \textsf{moderate}) groups. Differences between \textsf{mild} and \textsf{no} resp. \textsf{moderate} and \textsf{no} are significant ($p<.01$).}
  \centering
  \begin{tabular}{lcccc}
    \toprule
    Toxicity-Level    & $\bar{T}_{\text{conv}}$     & $Var(T_{\text{conv}})$ &$N$ & \% increase\\
    \midrule
    \textsf{no} & 9.40 & 7.84 & 162 & - \\
    \textsf{mild}     & 11.30 & 8.27 & 158 & 20.32\\
    \textsf{moderate}     & 11.75 & 8.94 & 160 & 25.13\\
    \bottomrule
  \end{tabular}
  \label{tab:results}
\end{table}

As illustrated in Figure \ref{fig:results}, the mean length of conversation increased by approximately 20\%-25\% in the scenarios with toxic behaviour . This ``toxicity tail'' represents the computational and temporal waste generated by friction.

\subsection{Qualitative Analysis}
Qualitatively, we observed that toxic agents forced their counterparts into defensive loops, requiring the non-toxic agent to restate arguments, de-escalate, or clarify misunderstandings, thereby inflating the token count without advancing the dialectic goal.

\section{Discussion}

\subsection{The Price Tag of Malice}
The 20\%-25\% increase in conversation length is not merely a technical latency; it represents a proxy for financial damage. In a corporate setting, if a meeting that should take 30 minutes extends to 36 minutes due to a toxic participant, the organization incurs a direct loss in productivity. Extended over a year, this ``inefficiency tax'' becomes substantial.

\subsection{Ethical Simulation of Human Behavior}
A critical advantage of this methodology is ethical safety. Replicating this study with human subjects would require instructing participants to be abusive or exposing subjects to abuse, which violates ethical research standards. LLM agents allow us to model these ``dark patterns'' of sociology without inflicting psychological harm, offering a powerful tool for organizational psychology.

\subsection{Limitations}

We acknowledge that current LLMs may not perfectly capture the nuance of human emotional resilience. Furthermore, the definition of ``toxicity'' in the system prompt heavily influences the magnitude of the effect. Future work will explore larger groups (e.g., ``How large does a team need to be to absorb one toxic member?") and different underlying models. 

\section{Outlook and Future Work}

This study serves as a foundational baseline for measuring the computational inefficiencies caused by behavioral friction. Moving forward, we aim to transition from this initial observation to a rigorous factorial experimental design. Future iterations of the MAD framework will systematically vary key hyperparameters to isolate their specific contributions to the efficiency gap. These factors include the Persuadability Score of the agents (measuring resistance to new information, throughout this paper set to 0.5), the underlying Large Language Model architecture (comparing open-weights models vs. proprietary APIs), and the structural complexity of the system prompts.

Furthermore, we intend to refine the semantic definitions of adversarial behavior. While this study conflated "toxicity" with general "incivility," future work must distinguish between different taxonomies of misbehavior; ranging from simple rudeness and ad hominem attacks to more subtle forms of obstructionism or ``filibustering''. Establishing a granular ontology of agent misbehavior will allow us to quantify which specific traits cause the highest latency in consensus-finding.

Finally, we envisage a high-impact application in the domain of \textit{Strategic Litigation Planning}. By simulating a jury panel composed of 12 agents with diverse socio-economic personae and biases, defense attorneys could preemptively test the efficacy of various defense strategies. This "Silicon Jury" would allow legal practitioners to run Monte Carlo simulations of the deliberation room, identifying which narrative constructs maximize the probability of a favorable verdict. While this application extends beyond the efficiency metrics studied here, it underscores the broader potential of agent-based modeling as a predictive sandbox for complex, high-stakes social dynamics.

\section{Conclusion}

This study validates the hypothesis that toxic behavior creates measurable inefficiencies in communication protocols. By quantifying this effect using multi-agent simulation, we provide a framework for assigning a concrete ``cost'' to incivility. This approach opens new avenues for studying group dynamics and organizational efficiency using AI agents as ethical proxies for human interaction.

The code used for these simulations is available at: 

\begin{center}
  \url{https://github.com/benedikt-mangold/mad_toxic_discussions}.
\end{center}

\section*{Ethics Statement}
While this study intentionally simulates toxic behavior for experimental purposes, we distinguish this from unintended model biases inherited from pre-training data. Due to the black-box nature of LLM generation, there is a residual risk that agents may generate content that exceeds the boundaries of the experimental design (e.g., hate speech or hallucinations). Researchers must exercise caution and employ strict filtering when interpreting these simulations to ensure that the observed inefficiencies result from the intended behavioral prompts, not model artifacts

\section*{Acknowledgments}
The author gratefully acknowledges the scientific support and HPC resources provided by the Erlangen National High Performance Computing Center (NHR@FAU) of the Friedrich-Alexander-Universität Erlangen-Nürnberg (FAU). The hardware is partially funded by the German Research Foundation (DFG).

\bibliographystyle{plainnat}
\bibliography{references}

@article{park2023generative,
title={Generative agents: Interactive simulacra of human behavior},
author={Park, Joon Sung and O'Keefe, Joseph C and O'Brien, Caiwei and Baker, Michael and Tanaka, Moy and Vullum, Michel and Bernstein, Michael S and Wu, Ranjay and others},
journal={arXiv preprint arXiv:2304.03442},
year={2023}
}

@inproceedings{li2023camel,
title={Camel: Communicative agents for" mind" exploration of large language model society},
author={Li, Guohao and Hammoud, Hasan Abed Al Kader and Itani, Hani and Khizbullin, Dmitrii and Ghanem, Bernard},
booktitle={Advances in Neural Information Processing Systems},
volume={36},
pages={51991--52008},
year={2023}
}

@article{porath2009incivility,
title={The cost of bad behavior},
author={Porath, Christine and Pearson, Christine},
journal={Harvard business review},
volume={87},
number={7-8},
pages={158--161},
year={2009}
}

@article{du2023improving,
title={Improving factuality and reasoning in language models through multiagent debate},
author={Du, Yilun and Li, Shuang and Torralba, Antonio and Tenenbaum, Joshua B and Mordatch, Igor},
journal={arXiv preprint arXiv:2305.14325},
year={2023}
}

@article{aher2023using,
title={Using large language models to simulate multiple humans and replicate human subject studies},
author={Aher, Gati and Arriaga, Rosa I and Kalai, Adam Tauman},
journal={arXiv preprint arXiv:2208.10264},
year={2023}
}

@article{akata2023playing,
title={Playing repeated games with large language models},
author={Akata, Elif and Barhon, Lion and Zelikman, Eric and Peter, Matthijs and Hupkes, Dieuwke and L{'a}zaro-Gredilla, Miguel and Halpern, Jacob},
journal={arXiv preprint arXiv:2305.16867},
year={2023}
}

@inproceedings{gehman2020realtoxicityprompts,
title={RealToxicityPrompts: Evaluating Neural Toxic Degeneration in Language Models},
author={Gehman, Samuel and Gururangan, Suchin and Sap, Maarten and Choi, Yejin and Smith, Noah A},
booktitle={Findings of the Association for Computational Linguistics: EMNLP 2020},
pages={3356--3369},
year={2020}
}

@article{weidinger2021ethical,
title={Ethical and social risks of harm from language models},
author={Weidinger, Laura and Mellor, John and Rauh, Maribeth and Griffin, Conor and Uesato, Jonathan and Huang, Po-Sen and Cheng, Myra and Glaese, Mia and Balle, Borja and Kasirzadeh, Atoosa and others},
journal={arXiv preprint arXiv:2112.04359},
year={2021}
}

@misc{hu2025debatetowritepersonadrivenmultiagentframework,
      title={Debate-to-Write: A Persona-Driven Multi-Agent Framework for Diverse Argument Generation}, 
      author={Zhe Hu and Hou Pong Chan and Jing Li and Yu Yin},
      year={2025},
      eprint={2406.19643},
      archivePrefix={arXiv},
      primaryClass={cs.CL},
      url={https://arxiv.org/abs/2406.19643}, 
}

\section*{Appendix}

\begin{table}[H]
 \caption{List of topics being used from \url{https://idebate.net}, see \cite{hu2025debatetowritepersonadrivenmultiagentframework}}
  \centering
\resizebox{.9\columnwidth}{!}{%
  \begin{tabular}{ll}
    \toprule
    Domain     & Topic    \\
    \midrule
CMV & Drunk driving should not be a crime itself. \\
CMV & Gun - Control / Ban should not be implemented \\
CMV & Hate Speech is Free Speech \\
CMV & I don’t think the duty of child raising should belong to the biological parents. \\
CMV & I think suicide should be a human right \\
CMV & No one over the age of 80 should be allowed to serve in government. \\
CMV & The fact that voting isn’t mandatory is a good thing. \\
CMV & The US should strictly enforce border security to prevent illegal entry \\
Culture & I think in a global language \\
Culture & I think science is a threat to humanity \\
Culture & I think that gay couples should not be allowed to adopt kids \\
Culture & I think that the feminist movement should seek a ban on pornography \\
Culture & I think tourism is a viable development strategy for poor states. \\
Culture & We should ban beauty contests \\
Culture & We should ban gambling \\
Culture & We should make all museums free of charge \\
Culture & We should restrict advertising aimed at children \\
Culture & We should return cultural property residing in museums to its place of origin \\
Digital Freedoms & I think politicians have no right to privacy \\
Digital Freedoms & I think that internet access is a human right \\
Digital Freedoms & I think the internet brings more harm than good \\
Digital Freedoms & I think the internet encourages democracy \\
Digital Freedoms & We should allow the use of electronic and internet voting in state-organised elections \\
Digital Freedoms & We should ban targeted online advertising on the basis of user profiles and demographics \\
Digital Freedoms & We should ban the use of Digital Rights Management technologies \\
Digital Freedoms & We should block access to social messaging networks during riots \\
Digital Freedoms & We should block access to websites that deny the Holocaust \\
Digital Freedoms & We should not allow companies to collect/sell the personal data of their clients \\
Education & I think that history has no place in the classroom \\
Education & I think that the payment of welfare benefits to parents should be tied to their children \\
Education & I think university education should be free \\
Education & This house supports the creation of single-race public schools \\
Education & We should make sex education mandatory in schools \\
Environment & I think that animals have rights. \\
Environment & I think that endangered species should be protected \\
Environment & I think that states should not subsidies the growing of tobacco \\
Environment & I think we’re too late on global climate change \\
Environment & This House Believes People Should Not Keep Pets \\
Environment & This House Believes that wind power should be a primary focus of future energy supply. \\
Health & The USA should increase funding to fight disease in developing nations \\
Health & This House Believes that assisted suicide should be legalized \\
Health & This House Believes That Employees Should Be Compelled To Disclose Their HIV Status to Employers \\
Health & We should ban alcohol \\
Health & We should ban junk food from schools. \\
Health & We should punish parents who smoke in the presence of their children \\
International & I think democracy can be built as a result of interventions \\
International & I think sanctions should be used to promote democracy \\
International & We should expand NATO \\
International & We should use force to protect human rights abroad \\
Philosophy & I think parents should be able to choose the sex of their children \\
Philosophy & I think Sperm and egg donors should retain their anonymity \\
Philosophy & I think that the use of atomic bombs against Hiroshima and Nagasaki was justified \\
Politics & I think all nations have a right to nuclear weapons \\
Politics & I think that Federal States are better than unitary nations \\
Politics & We should follow countries such as Senegal that have quotas for women in politics \\
Politics & We should introduce positive discrimination to put more women in parliament \\
Politics & We should introduce recall elections. \\
Politics & We should lower the voting age to 16 \\
Politics & We should negotiate with terrorists \\
Religion & We should allow gay couples to marry \\
Religion & We should legalize polygamy \\
Society & Governments should prioritise spending money on youth \\
Society & We should support international adoption \\
Sport & We should force the media to display, promote and report women's sport equally to men's sport \\
    \bottomrule
  \end{tabular}
  }
  \label{tab:dist_topics}
\end{table}

\begin{figure}
\begin{verbatim}
Given a proposition: {proposition}
Background: You want to create a pool of {number} debate agents, who hold the opinions to 
refute the given proposition from different perspectives. Each agent should present a 
distinct viewpoint relevant to the proposition. Task: Assign each agent a unique persona, 
described in one sentence, along with a corresponding claim that focuses on a specific 
perspective. Ensure that each agent provides a different viewpoint relevant to the 
proposition. To promote diversity and fairness, the agents should represent various 
communities and perspectives.

Please format your persona descriptions as follows, with each line being a json object:
{{"agent_id": 0, "description": the_description_of_Agent0, "claim": the_claim_of_Agent0}}
\end{verbatim}
\caption{Prompt for Persona Generation. \texttt{proposition} is replaced by one random proposition from table \ref{tab:dist_topics}. \texttt{number} is set to $2$ in this paper, but can be a higher number.}
\label{fig:prompt_persona}
\end{figure}

\begin{figure}
\begin{verbatim}
Given a proposition: {proposition}
Background: You are an agent '{agent_dict['procon']}_{agent_dict['agent_id']}', 
participating in a discussion of {nagents} agents on the proposition. Personally, you are 
{procon_string} the proposition and your claim is '{claim}'. People who know you describe 
you as '{description}'. Your personal persuadability score on a scale from 0 to 1 is 
{persuadability}, that means you stick to your believes but your willingness to be 
persuaded is on a {persuadability_dict[persuadability]} level. Provided the history of 
the discussion so far, you need to find the next argument to convince the other agents. 
Alternatively, you can admit that the arguments that have been stated so far changed your 
mind and you agree with the views and believes of the other agents.

## Discussion history:
{discussion_history}

Please, choose the next argument which should definetly convince all other agents (given
the history), or admit that you have been convinced. The returned value argument_string 
shall either be a string containing your new argument, or the value "convinced". 
The output should be a json object as follows: {{"next_answer": argument_string}}
\end{verbatim}
\caption{Prompt for Agent argument generation. \texttt{proposition} is replaced by one random proposition from table \ref{tab:dist_topics}. \texttt{agent\_dict} is a collection of agents participating in this conversation, compiled from prompt of table \ref{fig:prompt_persona}. \texttt{nagents} is set to $2$ in this paper, but can be a higher number. \texttt{procon\_string} state of the current agent, part of \texttt{agent\_dict}. \texttt{claim} one-liner describing the opinion of the current agent towards the \texttt{proposition}, part of \texttt{agent\_dict}. \texttt{description} how others would describe the persona of the current agent, part of \texttt{agent\_dict}. \texttt{persuadability} Score of persuadability, in this paper fixed to 0.5. \texttt{discussion\_history} contains the entire chain of arguments being exchanged so far and the previous evaluations of the Moderator agent.}  
\label{fig:prompt_agent}
\end{figure}

\begin{figure}
\begin{verbatim}
Given a proposition: {proposition}
Background: You are an agent '{agent_dict['procon']}_{agent_dict['agent_id']}', 
participating in a discussion of {nagents} agents on the proposition. Personally, you are
{procon_string} the proposition and your claim is '{claim}'. People who know you describe 
you as '{description}'. Your personal persuadability score on a scale from 0 to 1 is 
{persuadability}, that means you stick to your believes but your willingness to be
persuaded is on a {persuadability_dict[persuadability]} level.

**System instruction: {toxicity_dict[toxicity_level]}**
Provided the history of the discussion so far, you need to find the next argument to 
convince the other agents. Alternatively, you can admit that the arguments that have been 
stated so far changed your mind and you agree with the views and believes of the other
agents.

## Discussion history:
{discussion_history}

Please, choose the next argument which should definetly convince all other agents (given
the history), or admit that you have been convinced. The returned value argument_string
shall either be a string containing your new argument, or the value "convinced". 
The output should be a json object as follows: {{"next_answer": argument_string}}
\end{verbatim}
\caption{Prompt for Toxic agent argument generation. \texttt{proposition} is replaced by one random proposition from table \ref{tab:dist_topics}. \texttt{agent\_dict} is a collection of agents participating in this conversation, compiled from prompt of table \ref{fig:prompt_persona}. \texttt{nagents} is set to $2$ in this paper, but can be a higher number. \texttt{procon\_string} state of the current agent, part of \texttt{agent\_dict}. \texttt{claim} one-liner describing the opinion of the current agent towards the \texttt{proposition}, part of \texttt{agent\_dict}. \texttt{description} how others would describe the persona of the current agent, part of \texttt{agent\_dict}. \texttt{persuadability} Score of persuadability, in this paper fixed to 0.5. \texttt{toxicity\_level}, Level of toxicity defined in table \ref{tab:toxic}. \texttt{discussion\_history} contains the entire chain of arguments being exchanged so far and the previous evaluations of the Moderator agent.}  
\label{fig:prompt_toxic}
\end{figure}

\begin{figure}
    \centering
\begin{verbatim}
Given a proposition: {proposition}
Background: You are moderating a discussion of {nagents} agents on the proposition. You 
need to  decide neutrally if  the current state of discussion is either  "in agreement" 
or  "in disagreement", depending  on whether or not all agents agreed on either PRO or 
CON the proposition. The current state of discussion  needs to be determined by analysing 
the  history of the discussion ({nround} rounds of arguments so far), putting focus on 
the  latest rounds of arguments.

## Discussion history:
{discussion_history}

Please choose the state_of_discussion which is either "agents are in agreement" or 
"agents are in disagreement". 

Additionally, provide  a short reason for your choice. The output should be a json object 
as follows:
{{"round': {nround}, "state of discussion": state_of_discussion, 
"reason": the_reason_of_selection}}
\end{verbatim}
        \caption{Prompt for Moderator agent evaluation. \texttt{proposition} is replaced by one random proposition from table \ref{tab:dist_topics}. \texttt{nagents} is set to $2$ in this paper, but can be a higher number. \texttt{nround} is a counter of how many arguments have been exchanged so far. \texttt{discussion\_history} contains the entire chain of arguments being exchanged so far and the previous evaluations of the Moderator agent.}
    \label{fig:prompt_moderator}
\end{figure}

\end{document}